\documentclass[letterpaper, 10 pt, conference]{ieeeconf}  

\IEEEoverridecommandlockouts                              

\usepackage{amssymb}
\usepackage{makecell}
\usepackage{subcaption}
\usepackage{booktabs}
\usepackage{upgreek}
\usepackage{verbatim} 

\newcommand{\ie}{{\em i.e.}}
\newcommand{\eg}{{\em e.g.}}

\newcommand{\vs}{{\em vs}}
\newcommand{\Fig}[1]{Fig. \ref{fig:#1}}

\newcommand{\Sect}[1]{Sect. \ref{sec:#1}}

\newcommand{\fFrame}{\psi}

\usepackage{graphicx}
\graphicspath{{./figures/}}
\DeclareGraphicsExtensions{.png , .jpg, .eps}

\usepackage[usernames]{xcolor}

\title{\LARGE \bf
Is the Pedestrian going to Cross? Answering by 2D Pose Estimation
}

\author{Zhijie Fang and Antonio M. L\'opez
\thanks{$^{1}$Zhijie and Antonio are with the Dpt. of Computer Science at the Universitat Aut\`{o}noma de Barcelona (UAB), and with the Computer Vision Center (CVC) at the UAB. {\tt\small \{zfang, antonio\}@cvc.uab.es}}%
}

\begin{document}

\maketitle
\thispagestyle{empty}
\pagestyle{empty}


\begin{abstract}
Our recent work suggests that, thanks to nowadays powerful CNNs, image-based 2D pose estimation is a promising cue for determining pedestrian intentions such as \emph{crossing} the road in the path of the ego-vehicle, \emph{stopping} before entering the road, and \emph{starting} to walk or \emph{bending} towards the road. This statement is based on the results obtained on non-naturalistic sequences (Daimler dataset), {\ie} in sequences choreographed specifically for performing the study. Fortunately, a new publicly available dataset (JAAD) has appeared recently to allow developing methods for detecting pedestrian intentions in naturalistic driving conditions; more specifically, for addressing the relevant question \emph{is the pedestrian going to cross?} Accordingly, in this paper we use JAAD to assess the usefulness of 2D pose estimation for answering such a question. We combine CNN-based pedestrian detection, tracking and pose estimation to predict the crossing action from monocular images. Overall, the proposed pipeline provides new state-of-the-art results.  
\end{abstract}

\section{INTRODUCTION}

Even there is still room to improve pedestrian detection and tracking, the state-of-the-art is sufficiently mature \cite{Geronimo:2014, Zhang:2017, Wojke:2017} as to allow for increasingly focusing more on higher level tasks which are crucial in terms of (assisted or automated) driving safety and comfort. In particular, knowing the intention of a pedestrian to cross the road in front of the ego-vehicle, {\ie} before the pedestrian has actually entered the road, would allow the vehicle to warn the driver or automatically perform maneuvers which are smoother and more respectful with pedestrians; it even significantly reduces the chance of injury requiring hospitalization when a vehicle-to-pedestrian crash is not fully avoidable \cite{Meinecke:2003}. 

\begin{figure}
\centering
\includegraphics[width=\columnwidth]{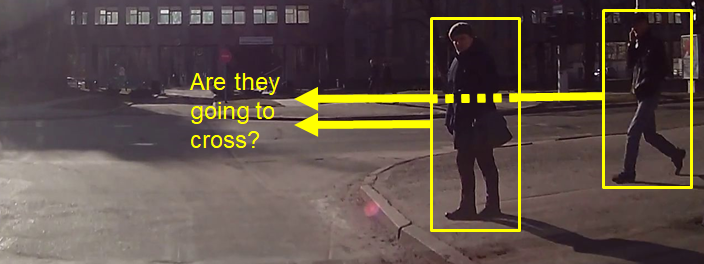}
\caption{Our focus: \emph{is the pedestrians going to cross?}}
\label{fig:cro-or-not-1frame}
\end{figure}

The idea can be illustrated with the support of \Fig{cro-or-not-1frame}. We can see two pedestrians, one apparently stopped near a curb and the other walking towards the same curb. Just looking at the location of the (yellow) bounding boxes (BBs) that frame these pedestrians, we would say that they are not in the path of the vehicle at the moment. However, we would like to know what is going to happen next: is the stopped pedestrian suddenly going to cross the road? is the walking pedestrian going to cross the road without stopping?; in the affirmative cases, the vehicle could start to slow down already for a safer maneuver, increasing the comfort of the passengers and the confidence of the pedestrians (especially relevant for autonomous vehicles).

Recently, we have addressed the crossing/not-crossing classification (C/NC) task by relying on image-based 2D pose estimation \cite{Fang:2017}. The proposed method shows state-of-the-art results and, in contrast to previous approaches (see \Sect{relatedWork}), it does not require information such as stereo, optical flow, or ego-motion compensation. As was common practice in the state-of-the-art literature, in \cite{Fang:2017} we used the only publicly available dataset for the C/NC task at that time, kindly released by Daimler researchers \cite{Schneider:2013}. While this dataset is a good starting point to challenge different ideas, it is composed of non-naturalistic sequences, {\ie} they show isolated pedestrians performing actions specifically choreographed for the C/NC task. Fortunately, a new dataset (\emph{Joint Attention for Autonomous Driving}--JAAD) has been publicly released recently \cite{Rasouli:2017}, which allows to address the C/NC task in naturalistic driving conditions. Accordingly, in this paper we present (see \Sect{method}) a pipeline consisting on a pedestrian detector, a multi-pedestrian tracker and a 2D pedestrian pose estimator, to obtain a per-pedestrian multi-frame feature set which allows to perform the C/NC task. Detector, Tracker and Pose Estimator are based on off-the-shelf CNN modules designed for such generic tasks, which we adapt here for our C/NC task. In this way, we can perform our experiments (see \Sect{experiments}) in the JAAD dataset. 

Therefore, with respect to \cite{Fang:2017}, we are facing a more challenging dataset for which using state-of-the-art pedestrian detection and tracking is mandatory. Note that for the dataset used in \cite{Fang:2017}, it was sufficient to rely on a simple HOG/Linear-SVM pedestrian detector and no tracking since the sequences only show single pedestrians under favorable illumination conditions. Moreover, since recently CNN-based features have been used to address the C/NC task in JAAD \cite{Rasouli:2017b}, we additionally compare our pose-estimation-based features with CNN-based ones. As we will see, the former clearly outperform the latter. Even more, as additional novelty we also report time-to-event (TTE) results in JAAD, which reinforce our argument about using pose estimation for detecting the crossing intentions of pedestrians. Overall, we think we are contributing with a new state-of-the-art baseline for JAAD, which is the only publicly available dataset at the moment acquired in naturalistic driving and containing ground truth annotations for the C/NC task.

\section{RELATED WORK}
\label{sec:relatedWork}

The C/NC task was initially taken as an explicit pedestrian path prediction problem; addressed by relying on pedestrian dynamic models for estimating pedestrian future location, speed and acceleration \cite{Schneider:2013, Keller:2014}. However, these models are difficult to adjust and for robustness require to rely on dense stereo data, dense optical flow and ego-motion compensation. Intuitively, methods like \cite{Keller:2014} implicitly try to predict how the silhouette of a tracked pedestrian evolves over time. In fact, \cite{Kohler:2015} uses a stereo-vision system and ego-motion compensation to explicitly assess the silhouette of the pedestrians (others rely on $360^\circ$ LIDAR \cite{Volz:2016}). Note that, while our method will be applied in JAAD because only relies on a monocular stream of images, these other methods cannot be applied due to the lack of stereo information and vehicle data for ego-motion compensation. 

On-board head and body orientation approximations have been also proposed to estimate pedestrian intentions, both from monocular \cite{Rehder:2014} and stereo \cite{Flohr:2015, Schulz:2015} images with ego-motion compensation. However, it is unclear how we actually can use these orientations to provide intention estimation. Moreover, the experiments reported in \cite{Schulz:2015} suggest that head detection is not useful for the C/NC task. 


These mentioned vision-based works relied on Daimler's dataset. By using an AlexNet-based CNN trained on JAAD, \cite{Rasouli:2017b} verified whether full body appearance improves the results on the C/NC task compared to analyzing only the sub-window containing either the head or the lower body. Conclusions were similar, {\ie} specifically focusing on legs or head does not seem to bring better performance. 

\begin{figure*}
\centering
\includegraphics[width=\textwidth]{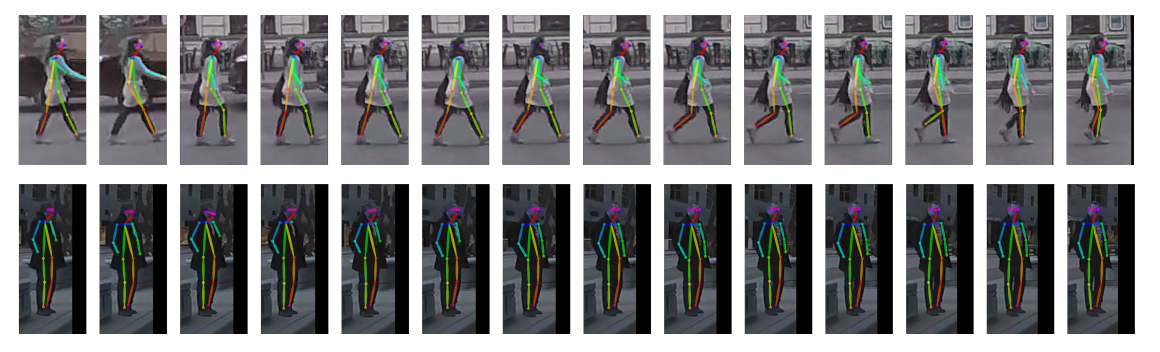}
\caption{Examples of 2D pose estimation by skeleton fitting. Top: pedestrian in side-view walking. Bottom: pedestrian standing still. From left to right we see 14 consecutive frames of two JAAD sequences, which roughly correspond to half a second.}
\label{fig:poseSequences}
\end{figure*}

\begin{figure}
\centering
\includegraphics[width=\columnwidth]{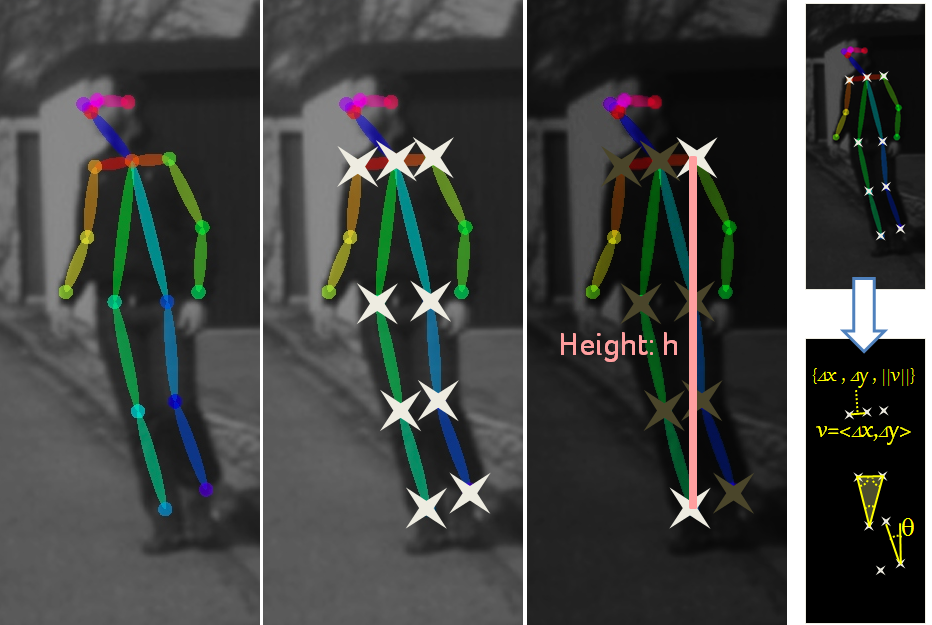}
\caption{Figure reproduced from \cite{Zhang:2017}. Skeleton fitting is based on 18 keypoints, distinguishing left and right \cite{Cao:2017}. We use the 9 keypoints highlighted with stars. The upper keypoint among those and the lower are used to compute height $h$, which is used as scaling factor for normalizing the keypoint coordinates. Then, using the normalized keypoints, different features based on relative angles and distances are computed as features. For instance, to the right we see several examples: (1) distance in the $x$ (column) and $y$ (row) axes and Euclidean distance between two keypoints ($\Delta x$, $\Delta y$, $\|v\|$); (2) angle between two keypoints ($\theta$); (3) the three angles of a triangle formed by three keypoints. After normalizing by $h$ these seven values, they become components of the feature vector $\fFrame_i$ of frame $i$. Computing similar values by taking into account all the keypoints we complete $\fFrame_i$.}
\label{fig:skeleton}
\end{figure}

In fact, in \cite{Schneemann:2016} it is concluded that \emph{a lack of information about the pedestrian's posture and body movement results in a delayed detection of the pedestrians changing their crossing intention}. In line with this suggestion, in \cite{Fang:2017} we relied on a state-of-the-art 2D pose estimation method that operates in still images \cite{Cao:2017}. In particular, following a sliding time-window approach, accumulating estimated pedestrian skeletons over-time (see \Fig{poseSequences}) and features on top of these skeletons (see \Fig{skeleton}), we obtained state-of-the-art results for the C/NC task in Daimler's dataset; which is remarkable since we only relied on a monocular stream of frames, but neither on stereo, nor on optical flow, nor on ego-motion compensation. In this paper, we augment our study to the more challenging JAAD dataset by complementing the 2D pose estimation with state-of-the-art pedestrian detection and tracking. Moreover, we compare the use of skeleton-based features with CNN-appearance-based ones as suggested in \cite{Gkioxari:2014} for the generic task of human action recognition. We will see how the former bring more accuracy than the latter. In addition, we also report TTE results.

\section{METHOD}
\label{sec:method}

In order to address the C/NC task we need to detect pedestrians, track them, adjust a skeleton for each one, frame by frame (see \Fig{poseSequences}), and apply a C/NC classifier for each pedestrian by relying on features defined on top of the respective skeleton (see \Fig{skeleton}). Accordingly, in this section we briefly describe the components used for detection, tracking, skeleton fitting (pose estimation) and C/NC classification.

\paragraph{Detection} For pedestrian detection we have fine-tuned a generic object detector based on the popular Faster R-CNN \cite{Ren:2015}. In particular, we have used the TensorFlow publicly available implementation described in \cite{Chen:2017}, based on a VGG16 CNN architecture. During the training stage of the C/NC classification pipeline, we have fine-tuned the model with JAAD training images.  

\paragraph{Tracking} Pedestrian tracking is addressed as a multiple object tracking-by-detection paradigm. A state-of-the-art tracker addressing this paradigm can be found in \cite{Wojke:2017}, which has associated publicly available code that we have used out-of-the-shelf. This tracker uses the following \emph{state} for a pedestrian detection: $(u, v, \lambda, h, \dot{x}, \dot{y}, \dot{\lambda}, \dot{h})$; where $(u, v)$ represents the central pixel of the BB, $\lambda$ is its aspect ratio, $h$ its height, while $\dot{x}$, $\dot{y}$, $\dot{\lambda}$, and $\dot{h}$ are the respective velocities. These state variables are updated according to Kalman filtering. For performing data association, it is used a cosine distance on top of CNN features (trained on a large-scale person re-identification dataset \cite{Zheng:2016}) which scores the degree of visual similarity between BB detections and predictions. A detection which does not have a high matching score with some prediction is pre-tracked; if the lack of matching holds during several consecutive frames (for JAAD we set 3 frames, {\ie} 0.1 seconds), the track is consolidated as corresponding to a new pedestrian. Predictions which do not have a high matching score with a new detection during several frames (for JAAD we set 30 frames, {\ie} 1 second) are considered as disappeared pedestrians (ended tracks). Note that this tracking process is purely image-based, no ego-motion compensation is required. 

\paragraph{Skeleton fitting (pose estimation)} Given the good results obtained in \cite{Fang:2017}, we apply the CNN-based pose estimation method proposed in \cite{Cao:2017}, which has publicly available code. This method can operate in still monocular images and has been trained on the \emph{Microsoft COCO 2016 keypoints dataset} \cite{Lin:2014}. It is supposed to perform both pedestrian detection and pose estimation. However, in our initial experiments with JAAD dataset, detection itself was not as good as Faster R-CNN. We think this is because, while we fine-tuned the later with JAAD images, we did not do the same for the pose estimation method since it would require annotations at pedestrian body level. Thus, what we do is to run the pose estimation only within the BBs predicted by the tracking system, obtaining in that way the desired skeletons (\Fig{poseSequences}).

\paragraph{C/NC classification} In \cite{Fang:2017} we extracted features from the fitted skeleton and use them as input to a classifier (SVM/Random Forest). \Fig{skeleton} shows that the fitted skeleton is based on 18 keypoints. We use the most stable 9 keypoints highlighted with a star, which correspond to the legs and the shoulders. These are highly relevant keypoints since the legs execute continue/start walking or stopping actions; while keypoints from shoulders and legs inform about global body orientation. From the selected keypoints we compute features. First, we perform a normalization of keypoint coordinates according to a factor $h$ proportional to the pedestrian height (\Fig{skeleton}). Then, different features (conveying redundant information) are computed by considering distances and relative angles between pairs of keypoints, as well as triangle angles induced by triplets of keypoints. In total we obtain 396 features. Since we concatenate the features collected during the last $T$ frames, our feature vector has dimension $396T$. In addition, for comparison purposes, as in the general action recognition literature \cite{Gkioxari:2014}, we also test the $fc6$ features provided by the Faster R-CNN at each pedestrian BB; a $4096T$ dimensional vector. Finally, since Random Forest (RF) directly provides a probability measure for a meaningful thresholding, we use it for performing the C/NC classification based on the selected features (skeleton or $fc6$ based ones). 


\section{EXPERIMENTS}
\label{sec:experiments}

\subsection{Dataset}

First publicly available dataset for research on detecting pedestrian intentions is from Daimler \cite{Schneider:2013}. It contains 68 short sequences (9,135 frames in total) acquired in non naturalistic conditions and shows a single pedestrian per video, where the pedestrian is forced to perform pre-determined actions. More recently, it has been publicly released the Joint Attention for Autonomous Driving (JAAD) dataset \cite{Rasouli:2017}, acquired in naturalistic conditions and annotated for detecting C/NC actions. It contains 346 videos (most of them 5-10 seconds long) recorded on-board with a monocular system, running at 30 FPS with a resolution of $1920\times1080$ pixels. Videos include both North America and Eastern Europe scenes. Overall, JAAD includes $\approx$~88,000 frames with 2,624 unique pedestrians labeled with $\approx$~390,000 BBs. Moreover, occlusion tags are provided for each BB. Where $\approx$~72,000 (18\%) BBs are tagged as partially occluded and $\approx$~46,000 (11\%) as heavily occluded. In addition, although we are not using it in this paper, JAAD contains also context information (traffic signs, street width, etc.) that we may use in further studies to complement purely pedestrian-based information. 

\subsection{Evaluation protocol}

In \cite{Rasouli:2017b}, JAAD was used for assessing a proposed C/NC method. However, it is not explained how the JAAD data was divided into training and testing, and the corresponding code is not available. Therefore, here we have followed a protocol that we think is reasonable and can be  reproduced. First of all, We take the first 250 videos of JAAD for training and the rest for testing. Moreover, pedestrians are labelled with many different actions which we have mapped to C/NC as follows. We term as C to the crossing labels of JAAD, as well as the labels in \{clear-path, moving-fast, moving-slow, slow-down, speed-up\} assigned to a pedestrian with lateral motion direction; the rest are denoted as NC.

\subsubsection{Training} 

In order to fine-tune the Faster R-CNN we consider all the training frames and basically follow the same settings than in \cite{Chen:2017}, but using \{8, 16, 32, 64\} as anchors and 2.5 as BB aspect ratio ({\ie} pedestrian oriented). For fine-tuning we perform 110,000 iterations (remind that an iteration consists of a batch of 256 regions from the same image, and that input images are vertically mirrored to double the number of training samples). Regarding learning rate, we start with 0.001 and decrease the value to 0.0001 after 80,000 iterations.    

In order to train the C/NC classifier we needed to rely on well seen pedestrians as well as balancing the number of samples of the C and NC classes. For achieving this goal, we only consider pedestrian training samples with a minimum BB width of 60 pixels and no occlusion. Moreover, for a tracked pedestrian, these conditions must hold over more than T frames, since we need to concatenate last T frames for the C/NC classification. Thus, from tracks longer than T frames we can obtain different training samples by applying a temporal sliding window of size T. 

For each tracked pedestrian, the C/NC label assigned to a generated sequence of length T corresponds to the label in the most recent frame ({\ie} the frame in which the C/NC decision must be taken). We set T=14 for JAAD ({\ie}, following \cite{Fang:2017}, a value roughly below 0.5 seconds). Note that, since we are in training time, here we are referring to the ground truth tracks provided in JAAD. For completeness, we also test the case T=1; meaning that we only train with the last frame of the same sequences used for the T=14 case. Overall, there are 8,677 sequences of length T=14 and NC label, while there are 36,253 with C label; thus, in the latter case we randomly take only 8,677 among those 36,253 possible. Accordingly, we fit the pose estimation-based skeleton and compute the C/NC features (\Fig{skeleton}) for 8,677 C and 8,677 NC samples (in a set of experiments for T=14, in another for T=1). These features are then used as input for the scikit-learn \cite{scikit-learn} function GridSearchCV; which is parameterized for training a Random Forest (RF) classifier using 5-fold cross-validation with the number of trees running on $\{100, 200, 300, 400, 500\}$ and maximum depth running on $\{7, 15, 21, 30\}$. The optimum RF in terms of accuracy corresponds to 400 trees and a maximum depth of 15, but we noted that all configurations provided very similar accuracy.

In order to compare skeleton-based features with CNN-based ones, we apply the following procedure. For all training images we run the VGG16 obtained during Faster R-CNN fine-tuning. Then, for the same tracks mentioned before, we replace the skeleton-based features by the fc6 layer features inside the tracked pedestrian BBs. Note that (\Sect{method}) we have $396T$ skeleton-based features and $4096T$ fc6-based ones for each sample reaching RF training. In terms of RF parameter optimization (number of trees and maximum depth), CNN-based features reported similar accuracy as was the case for skeleton-based ones. Therefore, we set the same parameters, {\ie} 400 trees and a maximum depth of 15. For the sake of completeness, we also combine skeleton and CNN-based features using the same RF parameters.

\subsubsection{Testing} 

In \cite{Rasouli:2017b} evaluations are single frame based (T=1 in our notation) and only pedestrians with an action label are considered (those mapped to C/NC) here. When designing our experiments, we have seen that not all pedestrians of JAAD are annotated with a BB. Therefore, when we run the detection and tracking modules, we are detecting and tracking some pedestrians which do not have the required ground truth information (BB, etc.). So, in order to follow a similar approach to \cite{Rasouli:2017b}, we do not consider these cases for quantitative evaluation. However, they are present in the qualitative evaluation ({\eg} see the videos provided as supplementary material). Overall, we ensure that T=1 and T=14 experiments are applied at the same tracked pedestrians at the same frames, so we perform a fair comparison. 

When detecting pedestrians with Faster R-CNN we use the default threshold $5\%$ and overlapping of $30\%$ for non-maximum suppression. For starting a new track, a pedestrian must be detected in 3 consecutive frames; while for ending a track there must be no new matched observations (detections) during 30 frames. For pose estimation (skeleton fitting) we use 3 scales; in particular, $\{1, (1-0.15), (1-0.15*2)\}$. For the C/NC classifier we use 0.5 as classification threshold.   

\begin{figure}[t!]
\centering
\includegraphics[width=\columnwidth]{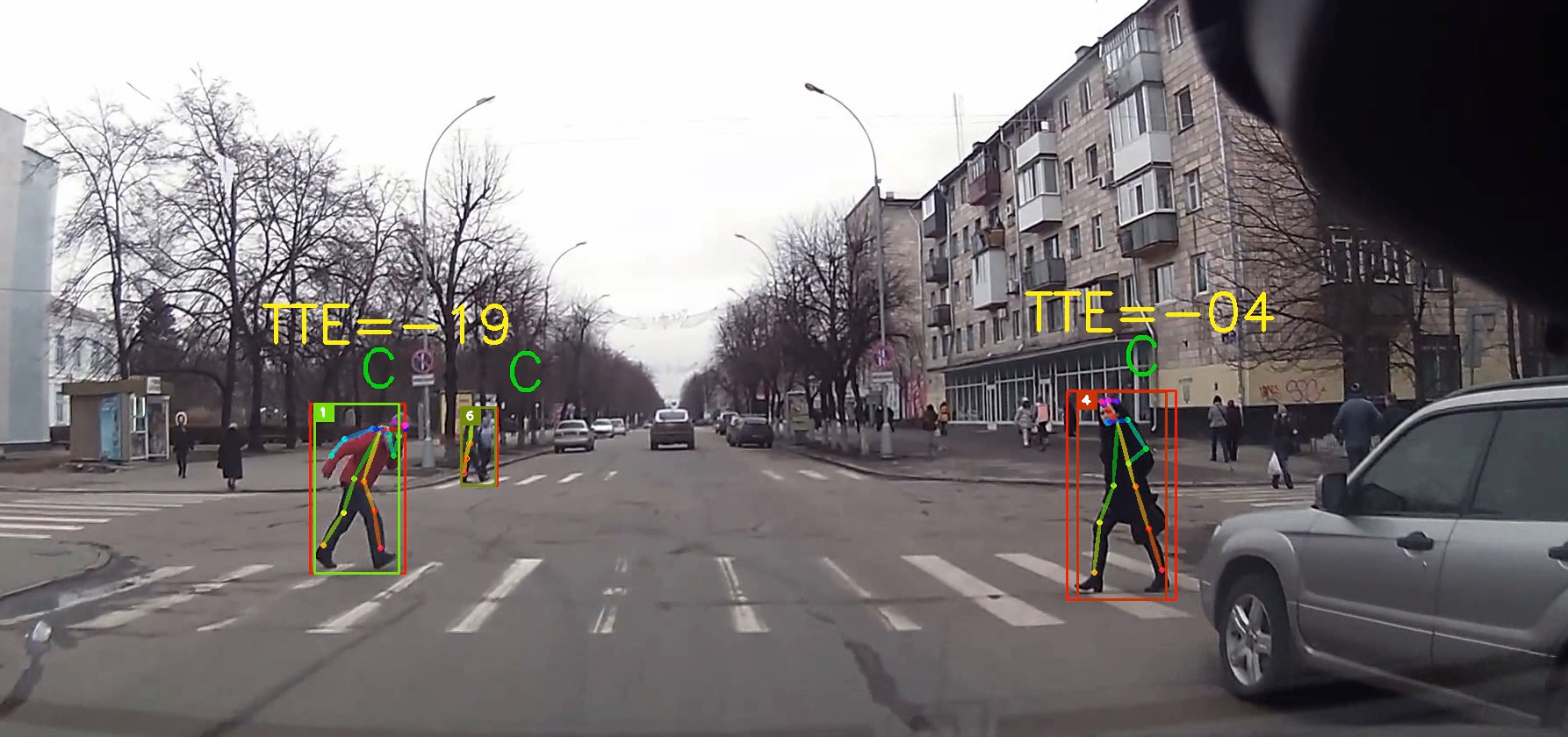}
\caption{Results of C/NC classification. The ground truth label is indicated with a "C" or a "NC"; when written in green color, it means that the prediction agrees with the ground truth, otherwise it would be written in red. Pedestrians are framed with two BBs: detection and tracling ones, the latter with the corresponding track ID. The estimated pedestrian skeleton is also shown. When annotated, time-to-event (TTE) is also shown in frame units. Negative TTE values mean that the even happened before this frame, while positive values indicate that it will happen after.}
\label{fig:res_c_nc_good}
\end{figure}

We assess accuracy according to the widespread definition $Acc = (TP+TN)/(P+N)$, where $P$ stands for total positives (here "C"), $N$ are the total negatives (here "NC"), and $TP$ and $TN$ the rightly classified positives and negatives (C and NC right classifications). According to the testing protocol we have defined, we found $P=17045$ and $N=5161$, therefore, $Acc$ could be bias towards "C" results. In order to avoid this, we select $P=N$ cases randomly. Thus, $Acc$ will be based on 10,322 testing decisions. 

In addition, similar to \cite{Fang:2017}, we are interested in providing time-to-event (TTE) results for the critical case of crossing (C). However, JAAD is not annotated for this. Then, we added the TTE information to 9 sequences we could describe as \emph{keep walking to cross}, and 14 more sequences we could describe as \emph{start walking to cross}. TTE = 0 is when the event of interest happens. Here we consider separately (a) pedestrians walking towards a curbside without stopping, just entering the road; and (b) pedestrians standing close to the curbside that start to walk entering the road. Positive TTE values correspond to frames before the event, negative values to frames after the event. \Fig{res_c_nc_good} shows a result example where we can see TTE values for different pedestrians that are correctly classified as crossing (the supplementary videos have more examples). With TTE we provide two different plots, \emph{intention probability} {\vs} TTE, and \emph{predictability} {\vs} TTE. With the former we can see how many frames we can anticipate the pedestrian action. Since there are several testing sequences per intention, mean and standard deviation are plotted. \emph{Predictability} plots show a normalized measurement of how feasible is to detect the action under consideration for each TTE value. Predictability zero indicates that we cannot detect the action, while predictability one means that we can.  

\begin{table}
\hspace*{-0.2cm}
\caption{Classification accuracy (Acc) in JAAD. SKLT stands for the use of our skeleton-based features, while CNN (fc6) are the features we take from a VGG16 fine-tuned in JAAD (see main text). We have included here the results reported in \cite{Rasouli:2017b}, where CNN features are based on a non-fine-tuned AlexNet and Context refer to features of the environment, not of the pedestrian itself (see main text).}
\centering

\begin{tabular}{ccc}
$\begin{array}{*{5}{c}}
\toprule
Method               &\textbf{T}	 & \textbf{features}  & \textbf{Acc} \\
\midrule
\cite{Rasouli:2017b} & 1             & CNN	              & 0.39 \\
\cite{Rasouli:2017b} & 1             & CNN \& Context     & 0.63 \\
\midrule
Ours                 & 1             & CNN (fc6)	      & 0.68 \\
Ours                 & 1             & SKLT	              & 0.80 \\
Ours                 & 1 & CNN (fc6) + SKLT   & 0.81 \\
\midrule
Ours                 & 14  & CNN (fc6)	      & 0.70 \\
Ours                 & 14  & SKLT	          & \textbf{0.88} \\
Ours                 & 14  & CNN (fc6) + SKLT & 0.87 \\
\bottomrule
\end{array} $
\end{tabular}
\label{tab:accuracy}
\end{table}

\subsubsection{Results} 

Table \ref{tab:accuracy} reports the accuracy results. In the sake of completeness, we have included those reported in \cite{Rasouli:2017b}; however, our results are not directly comparable since it is unclear which frames where used for training and which ones for testing. The paper mentions that heavily occluded pedestrians are not considered for testing. In our experiments we do not exclude pedestrians due to occlusion. Moreover, we also report TTE information. However, we still found interesting to include the results in \cite{Rasouli:2017b} since the paper is based on CNN features and T=1. In particular, the authors train a walking/standing classifier and another looking/not-looking (pedestrian-to-car) classifier, both classifiers are based on a modified AlexNet CNN. Actually, the classification score of these classifiers are not used for final C/NC decision. Instead, the fc8 layer of both are used as features to perform a final C/NC based on a Linear-SVM adjusted in such a CNN-based feature space. It is also proposed to add contextual information captured by a place-classification style AlexNet.  

\begin{figure}
\centering
\includegraphics[width=0.72\columnwidth]{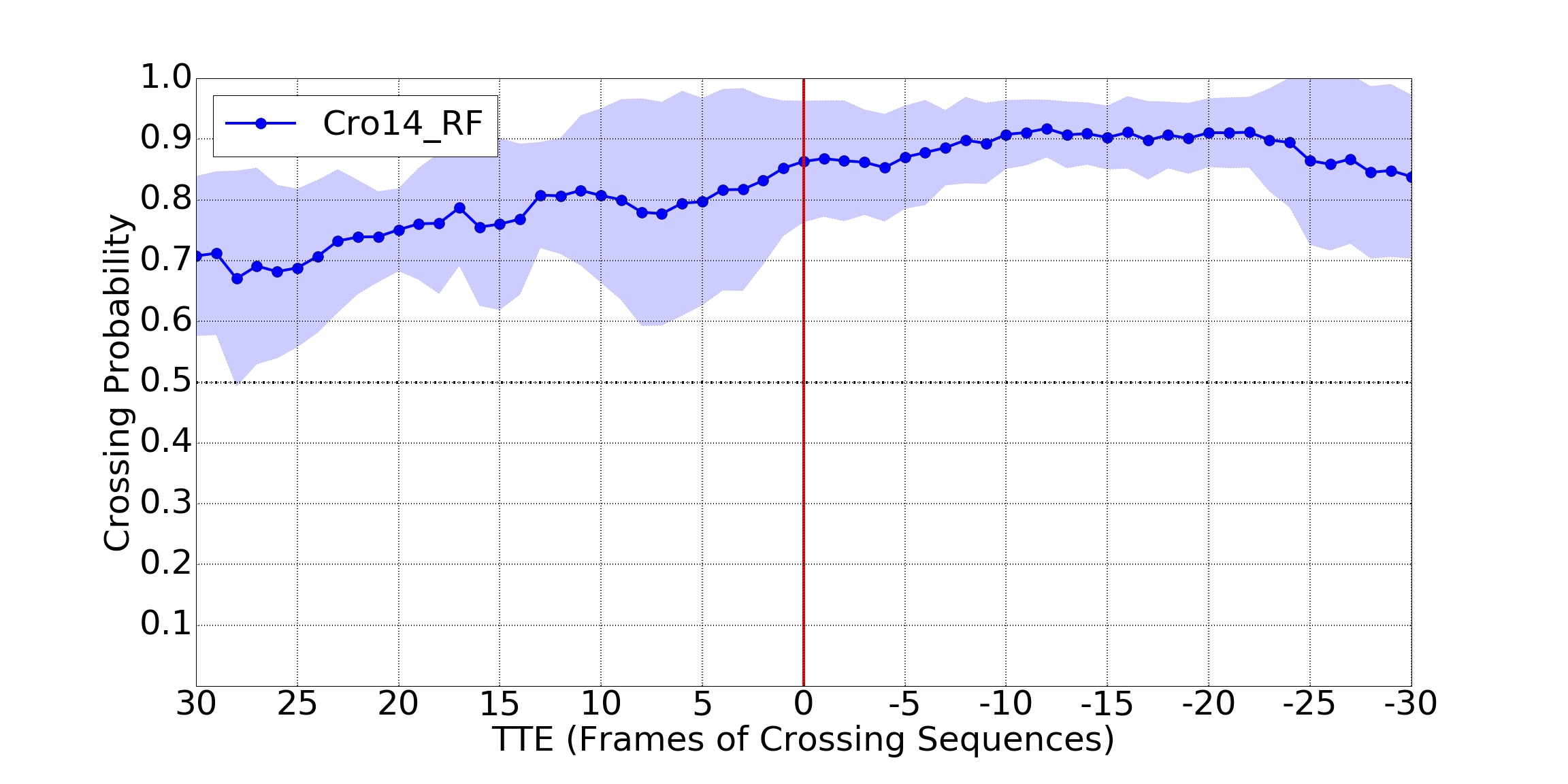}
\caption{\emph{Keep walking to cross}, T=14. Blue curve: mean over sequences; blue area: standard deviation.}
\label{fig:track_cro_prob}
\end{figure}

\begin{figure}
\centering
\includegraphics[width=0.72\columnwidth]{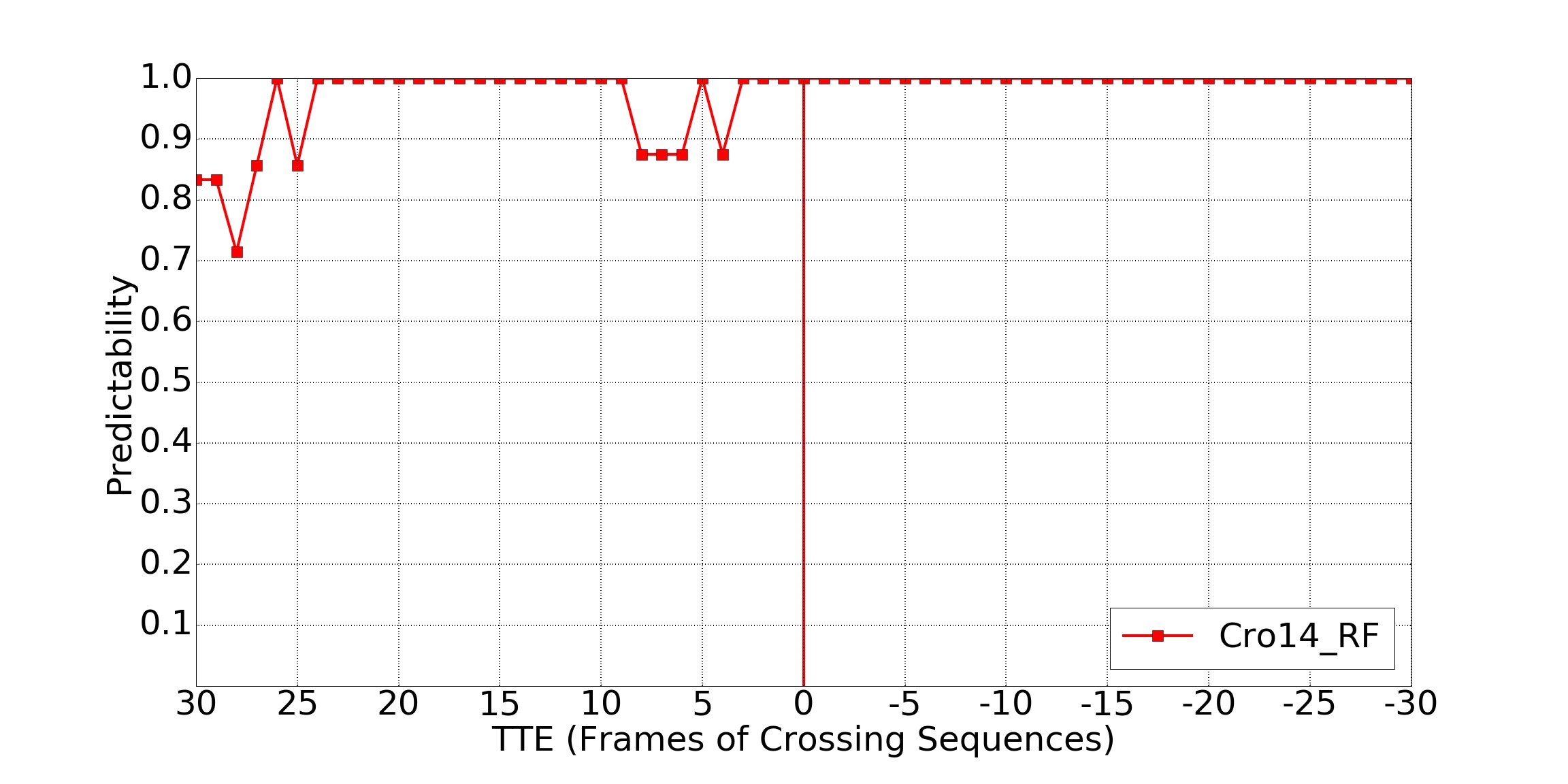}
\caption{\emph{Keep walking to cross}, T=14, prob. thr. = 0.5.}
\label{fig:track_cro_acc}
\end{figure}

\begin{figure}
\centering
\includegraphics[width=0.72\columnwidth]{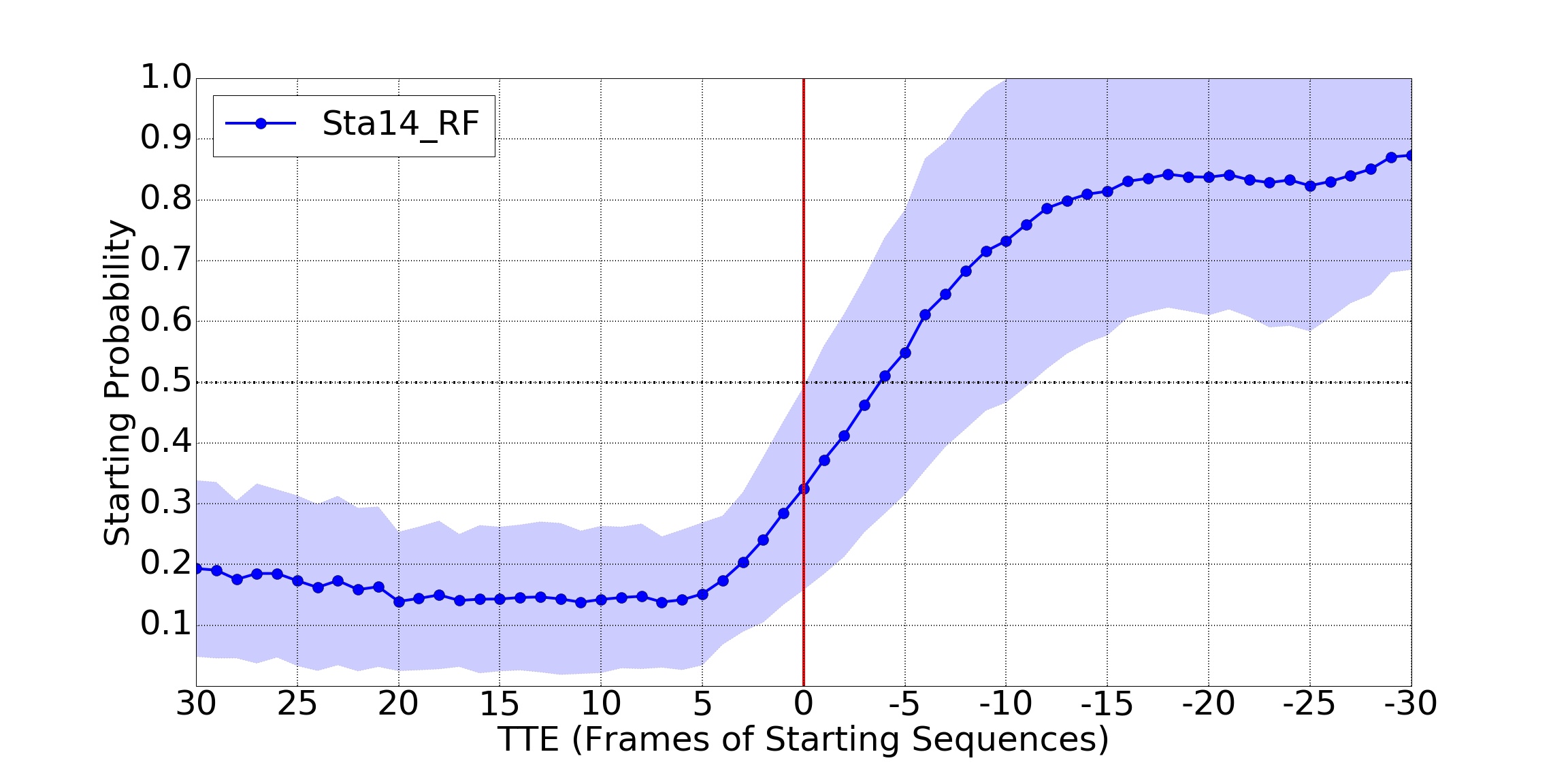}
\caption{\emph{Start crossing}, T=14.}
\label{fig:track_sta_prob}
\end{figure}

\begin{figure}
\centering
\includegraphics[width=0.72\columnwidth]{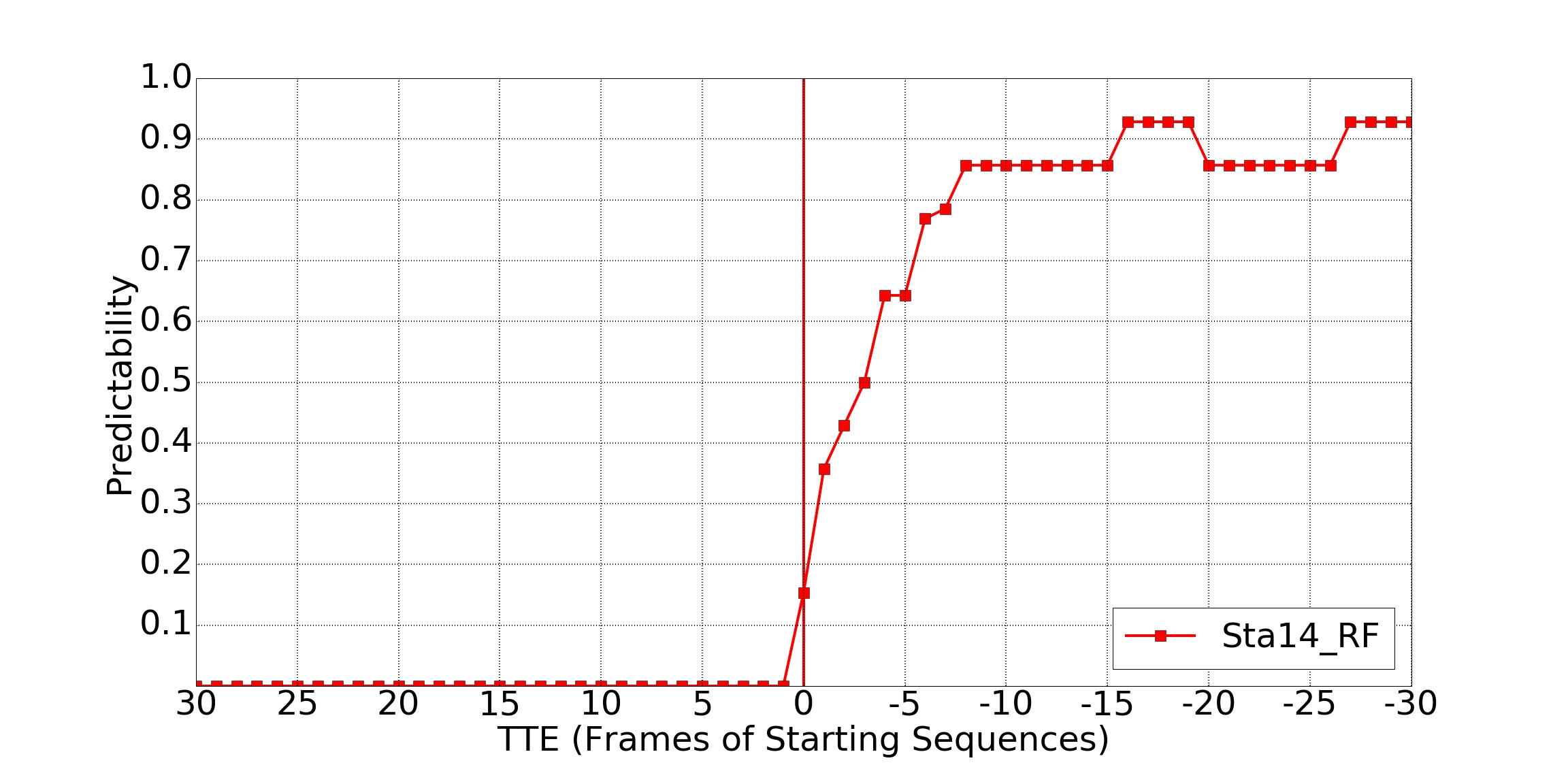}
\caption{\emph{Start crossing},  T=14, prob. thr. = 0.5.}
\label{fig:track_sta_acc}
\end{figure}

From Table \ref{tab:accuracy}, we can see that for a fixed T the features based on the skeleton of the pedestrian (SKLT) outperform those based on CNN fc6 layer. Combining SKLT and fc6 does not significantly improves accuracy of SKLT. We can see also that T=14 outperforms T=1, showing the convenience of integrating different frames. From \Fig{track_cro_prob} to \Fig{track_sta_acc}, we can see that the system is stable at predicting that a walking pedestrians will keep moving from a sidewalk and eventually crossing the curbside appearing in front of the vehicle. We can see also that we can predict (predictability$>$0.8) that a standing pedestrian will cross the curbside around 8 frames after he/she starts to move, which in JAAD is around 250ms. 

\begin{figure}
\centering
\includegraphics[width=\columnwidth]{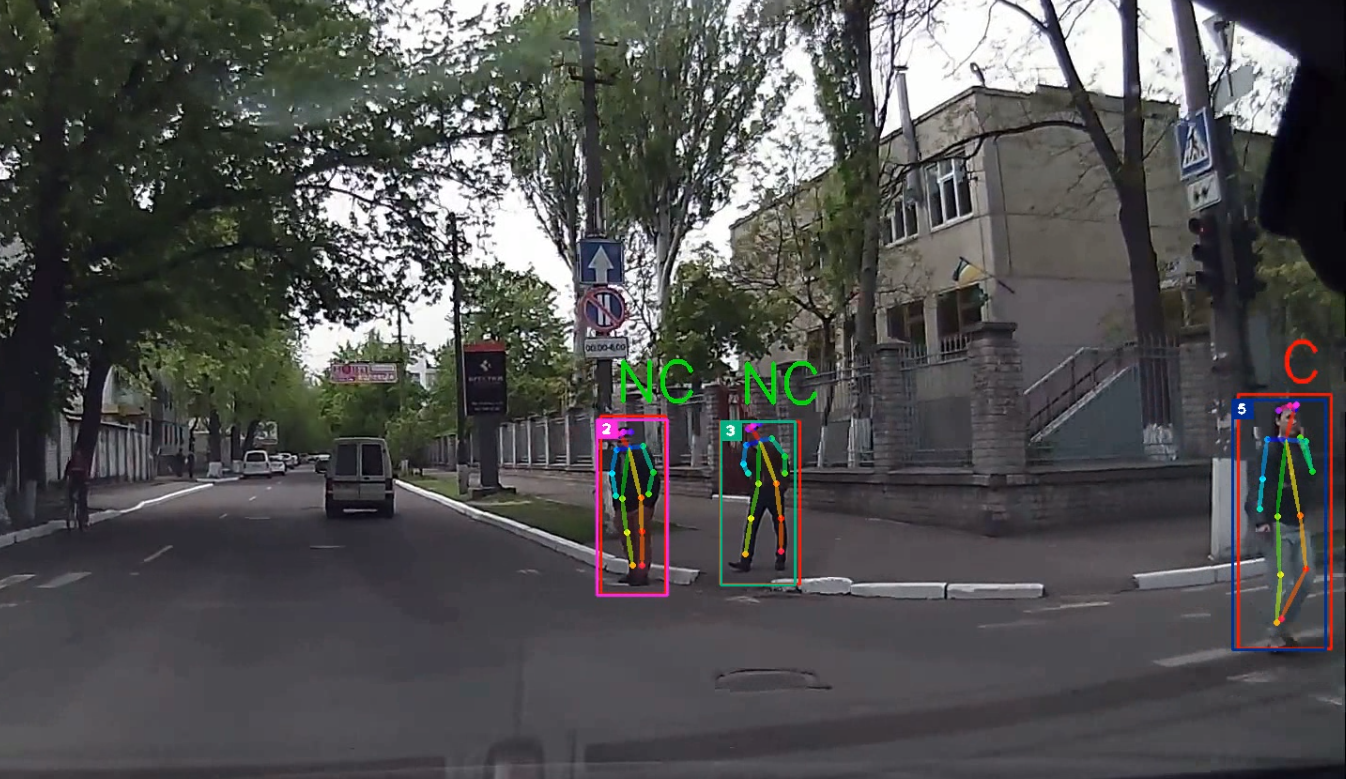}
\caption{Results of C/NC classification}
\label{fig:res_c_nc}
\end{figure}

\begin{figure}
\centering
\includegraphics[width=\columnwidth]{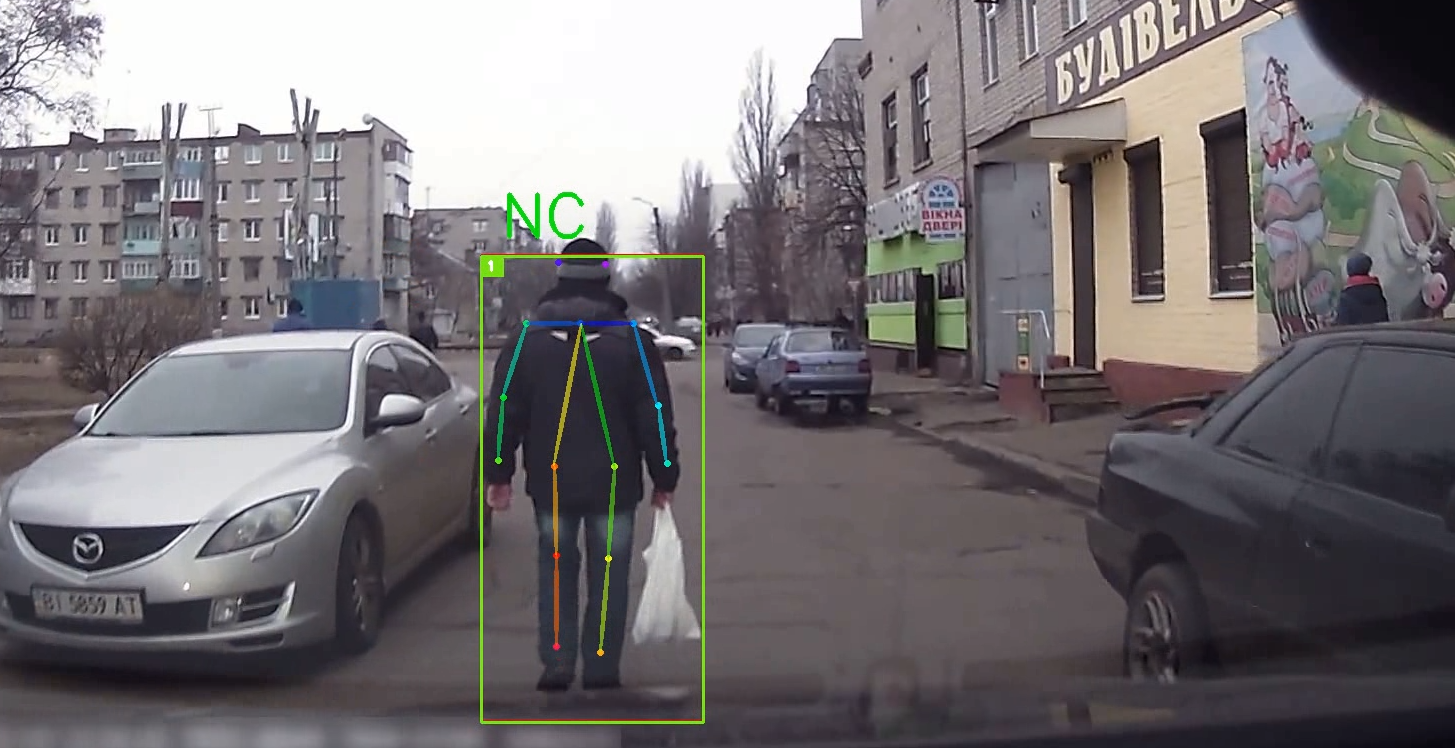}
\caption{Results of C/NC classification}
\label{fig:res_c_nc1}
\end{figure}

Looking in more detail to the results, we find situations that need to be taken into account as future work. For instance, in \Fig{res_c_nc} there is a "C" accounted as error (red). Indeed, the pedestrian is crossing the road, but not the one intersecting the path of the ego-vehicle. So in the evaluation it should be probably accounted as right. On the contrary, in \Fig{res_c_nc1} the system classifiers as "NC" a pedestrian which is not crossing the road, but in fact is walking along the road, in front of the car. Now this situation is accounted as right, but probably should be accounted as wrong. On the other hand, in this case we can just use location-based reasoning to know that the pedestrian is in a dangerous place, it is not a problem of predicting the action anymore (as the C/NC case). It is worth also mentioning that we have observed that walking in parallel to the car motion direction, tends to be properly classified as NC; however, more annotations are required to provide a reasonable quantitative analysis. Check our demo for more information (https://youtu.be/we4weU0NSGA).

\section{CONCLUSION}
\label{sec:conclusions}

In this paper we have evaluated a fully vision-based pipeline (detection, tracking and pose estimation) to address the pedestrian crossing/not-crossing problem, in naturalistic driving conditions (JAAD dataset). We show that integrating pedestrian pose based features along time, gives rise to a powerful crossing/not-crossing classifier. As to the best of our knowledge, at the moment this paper establishes the state-of-the-art results for the JAAD dataset.

%
%

\section*{ACKNOWLEDGMENT}
The authors want to acknowledge the Spanish project TIN2017-88709-R (Ministerio de Economia, Industria y Competitividad) and the Spanish DGT project SPIP2017-02237, the Generalitat de Catalunya CERCA Program and its ACCIO agency, as well as the China Scholarship Council, Grant No.201406150062. Our research is also kindly supported by NVIDIA Corporation in the form of GeForce GTX Titan X GPU hardware donations.





\end{document}